\documentclass[11pt]{article}
\usepackage{graphicx}
\usepackage{epstopdf} 
\usepackage{helvet}
\usepackage{setspace}
\usepackage{rotating}
\usepackage{booktabs}
\usepackage{lscape}
\usepackage[superscript]{cite}
\usepackage{caption}
\usepackage{longtable}
\usepackage{color}
\begin{document}
\thispagestyle{empty}
\fontfamily{phv}\selectfont{
{\LARGE \noindent The purpose of qualia:\\ What if human thinking is not (only) information processing?}
\\
\\
{\scriptsize Martin Korth$^{*}$\\
IVV NWZ, WWU Münster, Wilhelm-Klemm-Str. 10, 48149 Münster (Germany)\\
$^{*}$ Corresponding author email: dgd@uni-muenster.de\\
v1, 2022/11/21-22}\\ \\



\noindent
Despite recent breakthroughs in the field of artificial intelligence (AI) -- or more specifically machine learning (ML) algorithms for object recognition and natural language processing -- it seems to be the majority view that current AI approaches are still no real match for natural intelligence (NI).
More importantly, philosophers have collected a long catalogue of features which imply that NI works differently from current AI not only in a gradual sense, but in a more substantial way:
NI is closely related to consciousness, intentionality and experiential features like qualia (the subjective contents of mental states)
\cite{Chalmers}
and allows for understanding (e.g., taking insight into causal relationships instead of `blindly' relying on correlations), as well as aesthetical and ethical judgement beyond what we can put into (explicit or data-induced implicit) rules to program or train machines with.
Additionally, Psychologists find NI to range from unconscious psychological processes
to focused information processing, and from embodied and implicit cognition
to `true' agency and creativity.
NI thus seems to transcend any neurobiological functionalism
by operating on `bits of meaning' instead of information in the sense of data,
quite unlike both the `good old fashioned', symbolic AI of the past,
as well as the current wave of deep neural network based, `sub-symbolic' AI,
which both share the idea of thinking as (only) information processing:
In symbolic AI, the name explicitly references to its formal system based, i.e. essentially rule-based, nature,
but also sub-symbolic AI is (implicitly) rule-based, only now via globally parametrized, nested functions.
In the following I propose an alternative view of NI as information processing plus `bundle pushing',
discuss an example which illustrates how bundle pushing can cut information processing short,
and suggest first ideas for scientific experiments in neuro-biology and information theory as further investigations.
\\ \\

\noindent {\bf \fontfamily{phv}\selectfont{Materialism and thinking as information processing}}\\
\noindent 
The fact that we feel so strongly attached to the idea of thinking as information processing is most likely related to
the important role that materialism plays in our modern understanding of science:
If qualia, concepts, values etc. are understood as the outcome of material processes,
realized by changing constellations of material building blocks,
then information processing is surely the right way to conceptualize NI.
A theory for the formation of concepts (or more generally `meaning') from data,
in the most materialist sense as stable neural structures
accounting for regularities between actions and feedback,
is then correctly seen as the most important achievement to make.
But although materialism is a very powerful model indeed,
at the current stage of research into NI it does not seem likely that the above mentioned
complex philosophical and psychological phenomena
can easily be understood within that framework.
It therefore seems legitimate to see if alternatives to materialism could fare better,
with (at least) dualist, panpsychist and idealist positions available,
but none without its own conceptual issues:
While materialism has the above outlined emergence issue,
the dualist has to explain the interaction between her two worlds,
and while panpsychism (the assumption that on top of the material world
there are basic non-material building blocks) struggles with the de/combination
of minds from either `mind dust' or a `cosmic mind',
the idealist has to explain the `emanation' of the material world
from having only non-material building blocks.
I have elsewhere argued for a scientifically tenable objective (subject-independent) idealism,
\cite{MKscience}
which I believe allows for an intelligible interpretation of quantum theory,
\cite{MKquantum}
but the following should be suitable input for panpsychists and dualists, too.
\\ \\

\noindent {\bf \fontfamily{phv}\selectfont{An alternative view: Human thinking as information processing plus `bundle pushing'}}\\
\noindent
For the non-materialist, the core question regarding NI has to be,
how exactly we could understand human thinking as not (only) information processing
while staying true to modern science.
It is furthermore quite clear from neurobiology that large parts of human thinking can indeed
be understood as information processing:
This starts with the picking up of material signals, the conversion to neural activity,
the lower- and mid-level neural processing of senso-motoric data, but becomes somewhat
`blurry' at higher levels, where we can usually correlate
brain activity with mental activity, but not (yet?) in a strict sense.
\cite{Neuro2} 
It is here were the non-materialist can propose
that while brains serve as `anchors' for higher-level processes,
parts of these processes are not of material nature.
A first simple hypothesis would for instance be, that certain (e.g., cortical) brain regions
serve as a kind of `memory register' for mental entities:
Activity in a specific brain region invokes a specific mental entity and vice versa,
but the brain region is only a (material) handle for the actual (non-material) content.

Now, if the mental content is not fully defined by material constellation,
the problem then of course is how to define whatever needs to be added?
Not too surprisingly, science and philosophy have comparably few ideas to offer here.
In the context of idealism is seems somewhat natural to turn to `bundle theories' of objecthood in philosophical ontology,
which assume objects to be no more than the bundle of their properties, which in turn are traditionally taken to be universals (here: universal, non-material building blocks).
But because in this case objects with exactly the same bundle of universal qualities become essentially the same object,
such theories have a very non-materialist problem with the vanishing distinguishablity of indiscernable entities.
\cite{HS}
I have nevertheless argued elsewhere that this bug should actually be seen as a core feature,
as it allows for an intelligible interpretation of quantum theory
\cite{MKquantum}
within a scientifically tenable, bundle-theoretic view of objective idealism.
\cite{MKscience}
If we follow this ansatz (even if only due to the absence of good alternatives for our pupose here),
we would have to understand also the human mind as a bundle of universals,
anchored to a brain, which for the objective idealist would be just another sub-bundle of a whole person,
with certain special restrictions on the manipulation of the `brain-bundle' due to material causality.
As a dualist or panpsychist one could probably do with a simpler construction,
but could still understand the mind as a bundle of universals.
This `mind bundle' would serve as a `world map' for the individual,
charting (somewhat loosely related to time and space) the whole world of the agent,
who would also act on this map and not directly on the material world.
For such an individual, reality would not mean to understand an `ideal network of propositions' as critized by Heidegger,
but to `have a world', which would indeed supply local (subjective) context in universal, non-physical terms,
as required for human-like intelligence according to Dreyfus.\cite{Dreyfus}
\\ \\

\noindent {\bf \fontfamily{phv}\selectfont{Arguments for `bundle pushing'}}\\
\noindent
What have we gained at this point? We have formulated an alternative view
of human thinking as not only information processing, but closely related to it:
While at the bottom we have `abstractions' as signals from signals, i.e. information,
higher up we have (nested) sums of qualities as mental entities.
Conscious and most likely some `higher' parts of unconscious thinking are then `bundle pushing' (i.e. the shuffling of qualities) instead of information processing as material constellation pushing (i.e. `particle' shuffling).
This setup avoids the `materialist trap' of having to explain the emergence of meaning from data:
Information/data is transferred and shared via the material world in line with consistency rules for this part of the world,
but meaning is not transferred or directly shared, but instead generated by the individual
through the attachment of certain non-material building blocks to certain material signals or (higher up) other non-material building blocks,
i.e. the transfer of meaning would depend on shared (biologically/evolutionary, but also historically acquired) rules for the conversion of information to meaning.

Whether this alternative view is of any value should in my opinion be evaluated in a two-step process:
First we should check how the model fits to the list of relevant phenomena that long for an explanation.
Second, as with any (pre-)scientific model, we can try to derive predictions which lend themselves to further, also experimental investigation.
For the first step, we can acknowledge that the model does indeed (by design so to say) offer a route
to explain both the conscious, intentional, experiential and higher reasoning related features of NI,
as well as the unconscious, embodied, implicit cognition related ones.
With reference to learning, we can additionally think of mechanisms
that explain why when learning we need reference to existing knowledge, but still have to make the final move to a new `bit of meaning' by ourselves:
For an individual with a certain `heritage', a suitable set of material inputs will be strongly suggestive to make the right mental `move'.
And with reference to creativity, we can think of moves for bundle pushing
that allow for more complex intellectual steps than (sums of) inductive or deductive reasoning steps,
e.g., whole sub-bundles could be pushed according to less consistent or even `divergent' rules.
On the more philosophical side, the model would be well in line with the observation that our world is prone to scepticism,
also in the epistemological sense of critical idealism.
And on the more psychological side, the model would allow for a rich mental structure with strong sub-conscious/emotional forces;
direct access to our material body would be found at the sub-conscious level, making us prone to somatization.
Finally, the difference between (predominantly information-based) implicit vs (predominantly meaning-based) explicit execution of rules could explain observations like
Moravec's paradox and Kahneman's two modes of thinking.
\\ \\

\noindent {\bf \fontfamily{phv}\selectfont{The purpose of qualia: How bundle pushing can cut information processing short}}\\
\noindent
The most crucial question at this point probably is, how such a complex mind/brain construct could have developed
in line with our biological theory of evolution.
It is well established in neurobiology that the neural functioning of simple organisms can be fully explained as information processing,
\cite{Neuro2} 
so what would be an attainable(!) evolutionary advantage for more complex organisms to attach mental qualities to material signals?
(Or why else would life have ventured into the non-material after aeons in the material world only?)
Here again, like with quantum theory, the core bug of bundle theories might come to our help as a core feature of NI:
Imagine the effect for object recognition -- the first task at which modern AI has brought a break-through --, if let's say a bear is approaching.
In the model of symbolic AI, correctly identifying the bear as a bear would result
from repeated steps of information gathering and comparison to existing candidate objects.
(To avoid wrong early hits or flickering between results, certain thresholds for judgement would be helpful.)
With sub-symbolic AI the process would change from explicit rule following to implicit `cognition',
with the `rules' behind identifying a bear now being data-induced and opaquely distributed over several nodes of the neural network.
It seems quite clear that something like this sub-symbolic `thinking' needs to happen at the lower (information-processing) levels
of human thinking too, but unlike the former, the latter is for instance able to learn some things from extremely little data.

Our new view of NI would suggest that up to the generation of qualia, neural networks are a reasonable model for human thinking,
but everything afterwards would have to be understood as bundle pushing:
If the bear approaches, an up to now unidentified sub-bundle is generated and further extended
with additional qualities according to the additional information that is processed.
The fun thing now is that this addition of qualities does not only account for the collection of additional information,
but at the same time replaces the need for the repeated comparison of lists of properties
or the previous establishment of build-in implicit rules:
The more bear-like properties the bundle is collecting, the more identical -- in the literal sense! -- it becomes to the
existing `bear-bundle' in the world map of the individual,
because the properties that are pushed are universals, i.e. literally the same for all entities taking part in them.
Depending on the actual context, already at a rather early point
the existing and the newly adjusted relations can become more or less suddenly suggestive of a certain known bundle,
at which point the `bear-bundle' can simply switch into the current context,
as a side effect importing it's whole `bear-context', i.e. the sum of all other meaning attached to it.
Such a mechanism would thus not only explain why context, as the sum of pre-existing relations, plays a crucial role for human understanding,
but also why humans can be amazingly good at `zooming in' on content.
The model furthermore implies that if the correct object is identified, we would experience a `holistic' import of additional `every-day' knowledge,
which can indeed be observed for humans, but poses a giant challenge to realize for (sub-)symbolic AI.
\cite{Dreyfus}
(On the other hand, if things go wrong for us, they often go badly wrong: Illusions, hallucinations etc. are `fully real' for us until corrected.)

To answer the question at the beginning of this paragraph, we conclude that human thinking as bundle pushing might offer certain opportunities for cutting computational tasks short,
which could have been a evolutionary driving force for the development of complex non-/material mind/brain structures.
Similar `short-cuts' for informational tasks by recourse to universal qualities could of course also play a role for other complex mental activities,
like for instance natural language understanding. More generally understood, mental entities could allow for some kind of dimensionality reduction in complex optimization problems;
a simple movement in a low-dimensional `qualia space' could correspond to complex movements in higher dimensional spatiotemporal or informational spaces.
\\ \\

\noindent {\bf \fontfamily{phv}\selectfont{The next steps: First ideas for possible scientific investigations}}\\
\noindent
As mentioned above, the second step within our evaluation process would be to see whether we can derive predictions which lend themselves to further, if possible also experimental investigations.
It should be clear that actual experimental investigations would require a much more detailed theory of the underlying processes,
but the rough sketch of a model outlined above does already imply a number of questions to investigate:
Do `upper' brain circuits serve as `memory registers' only? If so, the length of 'entries' in the brain should be independent of the the actual content in the mind, etc.
Or can we influence the supposedly `arbitrary' (contingent) attachment between information and meaning, e.g. signals and qualia?
Furthermore, in our model, a mind would have to be build up in step with its brain; can we find evidence for such a complex integrated development?
(Later we might want to ask; does our model allow for helpful insights into psychopathology?)

Turning from neurobiology to information theory; can we proof that data can be cut short via meaning?
With this we move from `Does Mary learn something new?' (if she sees a color for the first time)
\cite{Mary1,Mary2}
to `How much does Mary learn?'
Does she learn a lot? An infinite amount? (Fully define red ...) Can we make use of this? To show the `supernatural' power of bundle pushing?
Probably not: We noted above that the transfer of meaning would be restricted to within one mind,
with material signaling always being restricted to finite information content, as material consistency means informational consistency.
But can we construct a human-solvable mental task that demonstrably surpasses the human brain's raw computing capacity?
Natural language understanding (as opposed to processing) might actually be such a task,
but is it possible to correctly specify the (context-dependent!) required computing power for a language task that would be complex enough?
And how to close loopholes relating human performance to evolutionary/biological/social adaption?

More far reaching implications of the proposed model might be open to investigation in the future:
Consciousness, qualia, mental causation, etc. would be no weird residues, but central features of any mind.
Accordingly, machines -- based on information processing only -- would never be able to fully achieve human intelligence.
But general AI  could nevertheless be possible, if we would be able to fully understand the coupled non-/material mind/brain development,
to make use of not only information processing artificial brains, but also bundle pushing artificial minds.
\\ \\

\noindent {\bf \fontfamily{phv}\selectfont{Conclusions}}\\
\noindent
In this manuscript I have put forward the claim
that alternatives to materialism allow us to propose models of human thinking and natural intelligence
beyond the current standard model of thinking as information processing only.
A simple object recognition example was discussed to illustrate how such models for information processing plus `bundle pushing' would work,
and in which ways bundle pushing could cut information processing short by recourse to universal qualities,
thereby implying that the initial `evolutionary purpose' of qualia could have been to short-cut computational tasks related to complex pattern recognition.
Further investigations into the new model are suggested, including first ideas for scientific experiments in neurobiology and information theory.
\\ \\





}
\end{document}